\documentclass[letterpaper, 10 pt, journal, twoside]{ieeetran}
\usepackage{subfigure}
\usepackage{amsmath,amsfonts}
\usepackage{algorithmic}
\usepackage{algorithm}
\usepackage{array}
\usepackage[caption=false,font=normalsize,labelfont=sf,textfont=sf]{subfig}
\usepackage{textcomp}
\usepackage{stfloats}
\usepackage{url}
\usepackage{verbatim}
\usepackage{graphicx}
\usepackage{cite}
\usepackage{multirow}
\usepackage{booktabs}
\usepackage{amssymb}
\usepackage{amsmath}

\usepackage{color, soul}
\soulregister\cite7
\soulregister\ref7
\soulregister\eqref7
\renewcommand\hl[1]{#1} 

\DeclareMathOperator{\atantwo}{atan2}

\newcommand{\etal}{\textit{ et al. }}
\begin{document}

\title{A Piecewise Monotonic Gait Phase Estimation Model for Controlling a Powered Transfemoral Prosthesis in Various Locomotion Modes}

\author{Xinxing Chen, Chuheng Chen, Yuxuan Wang, Bowen Yang, Teng Ma, Yuquan Leng, and Chenglong Fu
	\thanks{Manuscript received: February 24, 2022; Revised May 17, 2022; Accepted July 12, 2022.
		This paper was recommended for publication by Editor Pietro Valdastri upon evaluation of the Associate Editor
		and Reviewers’ comments. This work was supported by the National Natural Science Foundation of China [Grant U1913205, 62103180, and 52175272]; Guangdong Innovative and Entrepreneurial Research Team Program [Grant 2016ZT06G587]; the China Postdoctoral Science Foundation (2021M701577) ; the Science, Technology and Innovation Commission of Shenzhen Municipality [ZDSYS20200811143601004 and KYTDPT20181011104007]; the Stable Support Plan Program of Shenzhen Natural Science Fund [Grant 20200925174640002]; and Centers for Mechanical Engineering Research and Education at MIT and SUSTech.}
	\thanks{The authors are all with Shenzhen Key Laboratory of Biomimetic Robotics and Intelligent Systems, Shenzhen, 518055, China, and also with Guangdong Provincial Key Laboratory of Human-Augmentation and Rehabilitation Robotics in Universities, Southern University of Science and Technology, Shenzhen, 518055, China. Corresponding author: Chenglong Fu. Email: fucl@sustech.edu.cn.}
	\thanks{Ma is also with Department of Biomedical Engineering, National University of Singapore.}
	\thanks{Digital Object Identifier (DOI): see top of this page.}
}

\markboth{IEEE Robotics and Automation Letters. Preprint Version. Accepted July, 2022}
{Chen \MakeLowercase{\textit{et al.}}: A Piecewise Monotonic Gait Phase Estimation Model Controlling a Powered Prosthesis in Various Locomotion Modes} 


\maketitle

\begin{abstract}
Gait phase-based control is a trending research topic for walking-aid robots, especially robotic lower-limb prostheses. Gait phase estimation is a challenge for gait phase-based control. Previous researches used the integration or the differential of the human's thigh angle to estimate the gait phase, but accumulative measurement errors and noises can affect the estimation results. In this paper, a more robust gait phase estimation method is proposed using a unified form of piecewise monotonic gait phase-thigh angle models for various locomotion modes. The gait phase is estimated from only the thigh angle, which is a stable variable and avoids phase drifting. A Kalman filter-based smoother is designed to further suppress the mutations of the estimated gait phase. Based on the proposed gait phase estimation method, a gait phase-based joint angle tracking controller is designed for a transfemoral prosthesis. The proposed gait estimation method, the gait phase smoother, and the controller are evaluated through offline analysis on walking data in various locomotion modes. And the real-time performance of the gait phase-based controller is validated in an experiment on the transfemoral prosthesis.  
\end{abstract}


\section{Introduction}
\label{sec:1}
\IEEEPARstart{G}{ait} phase is a variable that describes the progress of a gait cycle. Compared with conventional finite-state control methods for powered lower-limb prosthesis \cite{4624583,6955797,lawson2012control,zhang2020subvision}, which have to tune several joint impedance controllers to mimic the damping and stiffness behaviors in separate gait stages, gait phase-based controllers determine the desired joint angles with unified models related to the estimated gait phase and require less manually-tuned hyper-parameters. A more significant superiority of the gait phase-based controllers is that they allow for semi-volitional control of the prosthesis because the users have the autonomy in adjusting their own gait phase. Therefore, gait phase-based control has been a trending research topic in the field of walking-aid robots \cite{rezazadeh2019phase,quintero2018continuous,rezazadeh2018phase,best2021phase,kang2019real,kang2021real}.

Gait phase-based control was initially proposed by Gregg\etal \cite{quintero2018continuous}. A human-inspired representation of the gait phase was utilized to parameterize the virtual constraint models of the joint angles of a powered prosthesis. The virtual constraints encoded the desired joint angles as functions of the monotonically increasing gait phase in each gait cycle.

A common challenge for gait phase-based control of walking-aid robots is gait phase estimation. Since the gait phase is highly correlated with a human's real-time pose during walking, accurate gait phase estimation can help determine the desired state of the robots through the virtual constraints in assistive walking. The gait phase is usually estimated from the human's lower-limb motion state measured by mechanical sensors, such as inertial measurement units (IMUs) \cite{cifuentes2014human,chen2021probability,chen2021multi} and pressure sensors \cite{kong2008smooth,martini2020pressure}. The thigh angle was one of the most widely-used variables for gait estimation, because it can robustly represent ipsilateral lower-limb motion patterns during unsteady walking \cite{7469796}.

Gregg\etal \cite{quintero2018continuous,8009448} estimated the gait phase by constructing a scaled and shifted phase plane of the thigh angle versus its integral. The phase angle calculated from the phase plane was then normalized and served as the gait phase. However, the computation of the phase angle presents a challenge for real-time control of the prosthesis. Besides, the integral of the thigh angle may introduce accumulative measurement errors in gait phase estimation. Hone\etal \cite{hong2021phase} also reported that the thigh angle profile and its integral were phase-shifted sinusoids and an additional phase correction method was required for accurate gait phase estimation. The gait phase can also be estimated from the pair of thigh angle and its angular velocity \cite{5152565,lee2021continuous,9350301}, but the measured angular velocity is prone to be noisy and make estimation inaccurate. \hl{The thigh angle can also be directly utilized to estimate the gait phase and it is reported to be more stable than the integral and the angular velocity\cite{rezazadeh2019phase,best2021phase,9795239}, but there was an obvious saturation of the estimated gait phase in the late swing phase in the previous works, which was caused by the retraction of the thigh before the heel-strike event and affected the estimation accuracy.}

\hl{Due to the above-mentioned challenges, most existing gait phase estimation methods were dedicatedly designed for a certain locomotion mode (i.e., terrain type), adopting various smoothing methods, phase plane shifting methods, and different definitions of the gait cycle to compensate for the measurement noise and the thigh retraction problem, yet a unified gait phase estimation method for various locomotion modes is rarely seen\cite{Zhang_2022} and has not been validated through online robotic experiments.}

To provide a solution for estimating gait phase in various locomotion modes, this paper proposes a unified piecewise monotonic model to describe the relationship between the gait phase and the human's thigh angle. \hl{Different from some previous gait phase estimation researches, the retraction stage is considered in the proposed model, because it is a common feature when traversing rough terrains\cite{4456910}, and can improve the stability of biped walk\cite{wisse2005swing}.} The inverse function of the model can then be utilized to estimate the gait phase from only the thigh angle. A Kalman filter-based smoother inspired by \cite{thatte2019robust} was designed to mitigate the mutations of the estimated gait phase. A powered transfemoral prosthesis is controlled to track the desired knee and ankle angles, which are determined by their virtual constraints and the estimated gait phase.

\hl{The main contributions of this paper are twofold:

(1) Towards gait phase estimation in various locomotion modes, a unified piecewise monotonic function of the thigh angle was proposed to estimate the gait phase, which can reduce the parameter tuning workload and make it easier to adapt the controller of the prosthesis to various terrain types.

(2) The thigh retraction stage, which was commonly observed when humans traverse through rough terrains, was taken into consideration to model a reverse motion of the swing leg before heel-strike. The added retraction stage helps improve the accuracy and adaptation of the proposed method in various locomotion modes and mitigates the potential saturation problem when using only the thigh angle.}

The proposed method is evaluated on a public dataset \cite{camargo2021comprehensive} and our collected lower-limb joint angle data of five able-bodied subjects. The results show that the proposed gait phase estimation method can achieve low estimation errors in various locomotion modes, including transition modes, which have rarely been analyzed in previous researches, especially for robotic prosthesis \cite{9364364}. Compared with other two gait phase estimation methods using the pair of $\phi$ and $\int \phi$ and using the pair of $\phi$ and $\dot \phi$, respectively, the RMSE of the smoothed gait phase estimated by the proposed method can reach $37.6\%$ lower than the other two methods.

The real-time performance of the proposed phase estimation method is validated through the experiments on a transfemoral prosthesis. The results demonstrate that the proposed gait phase estimation method and the controller can achieve good real-time gait phase estimation results and joint angle tracking performance.

The rest of the paper is organized as follows: Sec.~\ref{sec:II} describes the proposed method. The proposed models and phase smoother are evaluated offline with datasets in Sec.~\ref{sec:III}. Sec.~\ref{sec:III} also presents the walking experiment conducted on a powered transfemoral prosthesis and provides discussions and insights on the obtained results. Sec.~\ref{sec:IV} concludes the paper and presents outlooks on the gait phase-based control method.

\begin{figure*}[!t]
	\centering
	\includegraphics[width = 1.5\columnwidth]{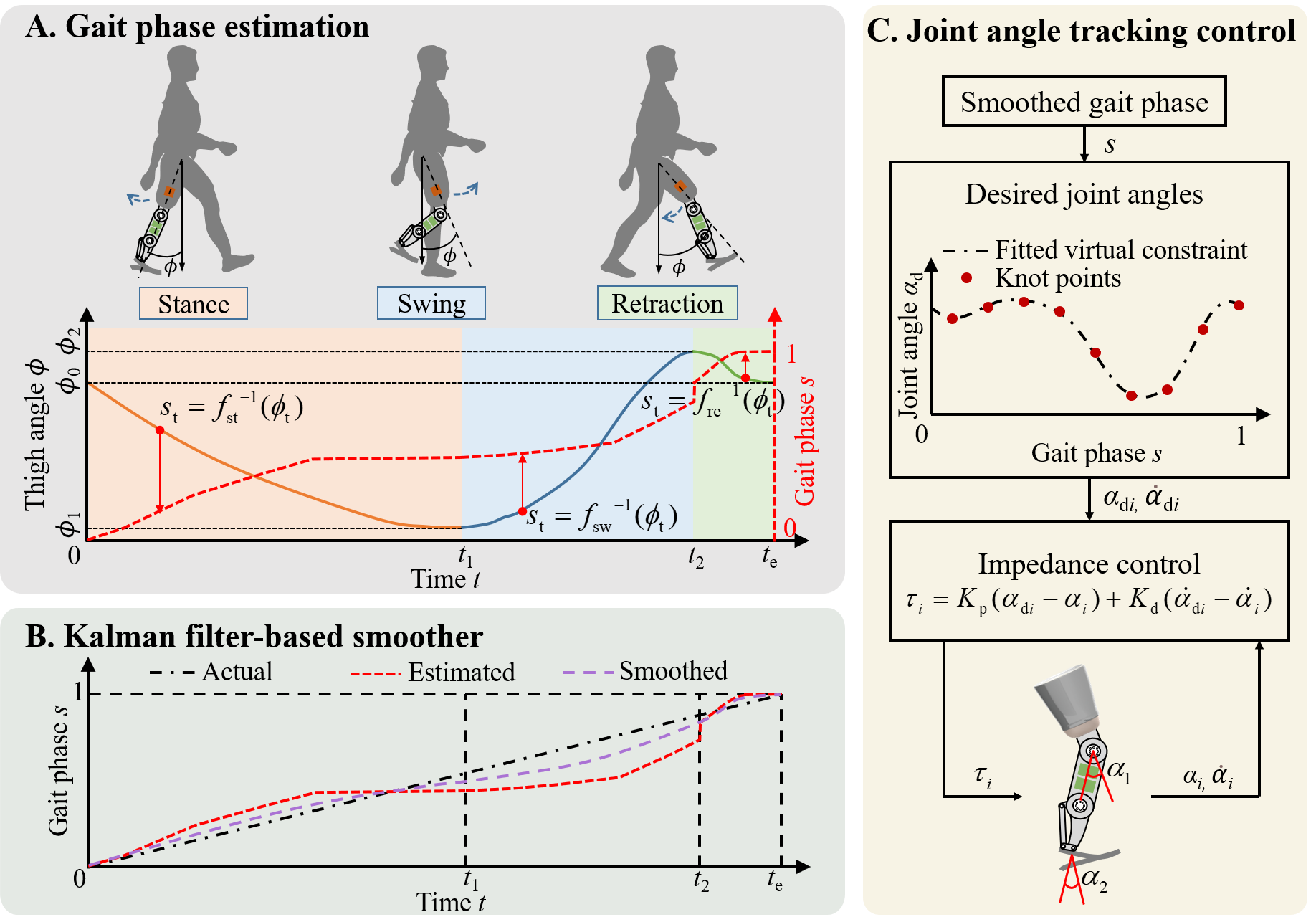}
	\caption{The proposed gait phase estimation method, the Kalman filter-based smoother, and the joint angle tracking controller for the powered transfemoral prosthesis. In (A), the gait cycle is divided into three stages, which are presented with different background colors. In each gait stage, the thigh rotates to the direction shown as the blue arrow. The thigh angle is illustrated with a colored solid curve. The gait phase (illustrated with the red dashed curve) is estimated from the thigh angle $\phi$ with the inverse function of the piecewise monotonic thigh angle-gait phase model. In (B), the black dashed-dotted line is a straight line increasing from 0 to 1, which is regarded as the actual gait phase in a gait cycle from time 0 to $t_e$. The red dashed curve is the estimated gait phase. The purple dashed curve is the gait phase smoothed by the Kalman filter-based smoother. In (C), the smoothed gait phase is used to calculate the desired joint angles. The prosthesis tracks the desired joint angles with a PD-like impedance controller.}
	\label{fig:main}
\end{figure*}

\section{Methods}
\label{sec:II}
In this section, the proposed gait phase estimation method based on a piecewise monotonic model was presented. A Kalman filter was designed to smooth the estimated gait phase. The desired joint angles of the powered prosthesis are determined by the estimated gait phase and the virtual constraints fitted from walking patterns of able-bodied subjects. Then a PD-like impedance controller is applied to the powered prosthesis to track the desired knee and ankle angles. Fig.~\ref{fig:main} presents an overview of the proposed gait phase estimation method and the controller for the prosthesis. Sec.~\ref{sec:II1} presents the mathematical derivation of the proposed phase variable. Sec.~\ref{sec:II2} shows how the Kalman filter-based smoother is designed. Sec.~\ref{sec:II3} introduces the virtual constraints between the estimated gait phase and the joint angles on various terrain types. The PD-like joint angle tracking controller of the powered prosthesis is also presented in Sec.~\ref{sec:II3}.

\subsection{Gait Phase Estimation}
\label{sec:II1}
As illustrated in Fig.~\ref{fig:main}, the thigh angle is not a monotonic variable during a gait cycle, which makes it hard to estimate the gait phase only from the thigh angle. Previous researches figured out a solution which introduced another variable (e.g., the integral or the differential of the thigh angle) to map the pair of two variables to the gait phase and solve the one-to-many problem. However, we have discussed in the Introduction that these supplementary variables are not stable in non-rhythmic walking and can bring in gait phase estimation drifting. In spite of this, the thigh angle itself is stable because it is directly measured from the human's pose with much fewer drifting and noise issues.

In this paper, in order to estimate the gait phase from only the non-monotonic thigh angle, an intuitive solution is to divide the gait cycle into a few stages to make the thigh angle monotonic in each stage. As shown in Fig.~\ref{fig:main}, a gait cycle can be divided into three stages: stance, swing and retraction. In the stance stage, the thigh rotates backwards around the hip joint, thus the thigh angle $\phi$ decreases monotonically from $\phi_0$ to $\phi_1$; once the swing stage starts, the thigh angle increases monotonically from $\phi_1$ to $\phi_2$; in the retraction stage, the thigh rotates backwards again until the heel-strike event, and the thigh angle decreases from  $\phi_2$ to $\phi_0$. Since the thigh angle is piecewise monotonic, as long as the gait stage can be divided accurately, the gait phase $s$ can be mapped to a certain thigh angle $\phi$.

In this paper, a set of biased sigmoid functions serve as the mapping functions from $s$ to $\phi$:
\begin{equation}
\phi = f_n(s) = \frac{h_n}{1+e^{-k_n\left(s-s_{0n}\right)}}+b_n, n\in\{\text{st, sw, re}\},
\label{eq:model}
\end{equation}
where $\text{st, sw, re}$ refer to the stance stage, the swing stage, and the retraction stage, respectively. $h_n, k_n, s_{0n}$ and $b_n$ are hyperparameters related to the shape of the functions and the optimal hyperparameters $\left(h_n, k_n, s_{0n}, b_n\right)_*$ can be determined by minimizing the gap between the fitted and the measured thigh angle curve:
\begin{equation}
\left(h_n, k_n, s_{0n}, b_n\right)_*=\mathop{\arg\min}\limits_{\left(h_n, k_n, s_{0n}, b_n\right)} \sum_{t}\left(f(s_t)-\phi_{t}\right)^{2}.
\label{eq:fitting}
\end{equation}
This set of functions are selected because they fulfill the following requirements for the mapping functions:

(1) The mapping functions should be monotonic in each gait stage: For any two gait phases $s_i$ and $s_j$ in the same gait stage $n$,
\begin{align}
\dot f_n(s_i)\cdot\dot f_n(s_j) &= \frac{k_n^{2} h_n^{2} e^{-k_n\left(s_{i}-s_{0n}\right)} e^{-k_n\left(s_{j}-s_{0n}\right)}}{\left(1+e^{-k_n\left(s_{i}-s_{0n}\right)}\right)^{2}\left(1+e^{-k_n\left(s_{j}-s_{0n}\right)}\right)^{2}}\nonumber\\&\geqslant 0, n\in\{\text{st, sw, re}\},
\end{align}
therefore, the selected functions are piecewise monotonic.

(2) The inverse function of the mapping functions should exist, so that the gait phase can be estimated from the thigh angle: The inverse function of the selected functions can be formulated as:
\begin{align}
\hat s=f_n^{-1}(\phi)=&s_{0n}-\frac{1}{k_n} \ln \left(\frac{h_n}{\phi-b_n}-1\right), \nonumber\\	
&\phi \in(b_n, b_n+h_n), n\in\{\text{st, sw, re}\},
\label{eq:estimate}
\end{align}
where $\hat s$ is the estimated gait phase.

(3) The inverse function in each gait stage should be continuous to avoid mutations of the estimated gait phase: The inverse functions are derivable in each gait stage, and the derivative functions can be expressed as Eq.~\eqref{eq:d}, therefore they are continuous.
\begin{align}
\dot f_n^{-1}(\phi)=&\frac{h_n}{k_n(b_n+h_n-\phi)(\phi-b_n)},\nonumber\\
&\phi \in(b_n, b_n+h_n), n\in\{\text{st, sw, re}\}.
\label{eq:d}
\end{align}

With the above piecewise monotonic gait phase-thigh angle model, the gait phase can be estimated from Eq.~\eqref{eq:estimate}, as illustrated by the red dashed curve in Fig.~\ref{fig:main}(a).

\subsection{Kalman Filter-based Smoother}
\label{sec:II2}
The gait phase estimated by Eq.~\eqref{eq:estimate} can be noisy due to thigh angle measurement noises and mutations may occur when gait stages switch (as illustrated in Fig.~\ref{fig:main}(b)) because the thigh angle curve in each gait cycle is not exactly the same as the fitted gait phase-thigh angle model in Eq.~\eqref{eq:model} and \eqref{eq:fitting}.

To mitigate the above issues, a Kalman filter-based smoother is designed to filter the noisy estimated phase by assuming that gait phase changing velocity does not vary largely in each gait cycle. The smoothed gait phase $\tilde s_{t}$ at time step $t$ can be calculated from estimated gait phase $\hat s_{t-1}$ and the phase changing velocity $ \dot{\hat{s}}_{t-1}$:

\begin{equation}
\left\{\begin{array}{l}
\boldsymbol{x}_t=\left[\begin{array}{l}
\tilde s_{t} \\
\dot{\tilde{s}}_{t}
\end{array}\right]=\boldsymbol{A}\boldsymbol{x}_{t-1}+\boldsymbol{B}_{t-1}\\
\boldsymbol{y}_{t-1}=\left[\begin{array}{l}
\hat s_{t-1} \\
\dot{\hat{s}}_{t-1}
\end{array}\right]= \boldsymbol{C}\boldsymbol{x}_{t-1}+\boldsymbol{D}_{t-1}
\end{array}\right.,
\end{equation}
where $\boldsymbol{x}_t$ is the smoothed state vector at time step $t$, $\boldsymbol{y}_{t-1}$ is the state vector estimated  by Eq.~\eqref{eq:estimate} at time step $t-1$.
$\boldsymbol{A}=\left[\begin{array}{cc}
1 & \delta t \\
0 & 1
\end{array}\right]$
is the state transition matrix, 
$\boldsymbol{C}=\left[\begin{array}{cc}
1 & 0 \\
0 & 1
\end{array}\right]$
is the observation matrix, $\boldsymbol{B} \sim N(0,\boldsymbol{Q})$ and $\boldsymbol{D} \sim N(0,\boldsymbol{R})$ are the process noises and measurement errors, respectively, where $\boldsymbol{Q}=\left[\begin{array}{cc}
0 & 0 \\
0 & {\sigma_B}^2
\end{array}\right]$ and $\boldsymbol{R} = \left[\begin{array}{cc}
{\sigma_{D1}}^2 & 0 \\
0 & {\sigma_{D2}}^2
\end{array}\right]$ are the covariance matrices. Since the gait phase changing velocity $\dot{\tilde s}$ is expected to be stable, $\sigma_B$ is set to be a small positive value in this paper, while $\sigma_{D1}$ and $\sigma_{D2}$ are determined based on experience.

A Kalman filter is utilized to update the above matrices through the following equations:
\begin{align}
\boldsymbol{x}_{t}&=\boldsymbol{A} \boldsymbol{x}_{t-1}, \\
\boldsymbol{P}_{t}&=\boldsymbol{A} \boldsymbol{P}_{t-1} \boldsymbol{A}^\mathsf{T}+\boldsymbol{Q},\\
\boldsymbol{K}_{t}&=\boldsymbol{P}_{t} \boldsymbol{C}^\mathsf{T}\left(\boldsymbol{C} \boldsymbol{P}_{t} \boldsymbol{C}^\mathsf{T}+\boldsymbol{R}\right)^{-1}, \\
\boldsymbol{x}_{t}&=\boldsymbol{x}_{t}+\boldsymbol{K}_{t}\left(\boldsymbol{y}_{t}-\boldsymbol{C} \boldsymbol{x}_{t}\right), \\
\boldsymbol{P}_{t}&=\left(\boldsymbol{I}-\boldsymbol{K}_{t} \boldsymbol{C}\right) \boldsymbol{P}_{t},
\end{align}
where $\boldsymbol{P_t}$ is the covariance matrix of the state, which indicates the uncertainty of the state $\boldsymbol{x}_t$, and $\boldsymbol{K}$ is the Kalman Gain.

\subsection{Joint Angle Tracking Control}
\label{sec:II3}
The models between the gait phase and the desired joint angles (i.e., the virtual constraints of the joint angles) can be fitted with collected walking data. In this paper, the virtual constraints $\alpha_{\text{d}i} = g_i(\tilde s),i \in \{1,2\}$ are built by fitting the actual joint angle values at all the knot points in a gait cycle to a cubic spline interpolation of the gait phase, where $i = 1$ is for the knee and $2$ is for the ankle, and $\alpha_{\text{d}i}$ is the desired joint angle.

Once the smoothed gait phase $\tilde s$ was obtained through the Kalman filter-based smoother, the desired knee angle $\alpha_{\text{d}1}$ and the ankle angle $\alpha_{\text{d}2}$ for the powered prosthesis can be determined. The desired joint angle and angular velocity then serve as the inputs of a PD-like impedance controller:
\begin{equation}
\tau_{i}=K_{\text{p}i}\left(\alpha_{\mathrm{d} i}-\alpha_{i}\right)+K_{\mathrm{di}}\left(\dot{\alpha}_{\mathrm{d} i}-\dot{\alpha}_{i}\right), i \in \{1,2\},
\label{eq:controller}
\end{equation}
where $K_{\text{p}i}$ and $K_{\text{d}i}$ are adjustable parameters, $\alpha_i$ and $\dot \alpha_i$ are the measured joint angle and angular velocity of the powered prosthesis. $\tau_{i}$ represents the current sent to the $i$th joint motor. By tuning $K_{\text{p}i}$ and $K_{\text{d}i}$, the powered prosthesis can track the desired joint angles accurately while performing human leg-like stiffness and damping behaviors.

\section{Performance Evaluation}
\label{sec:III}
The proposed method was evaluated on both a public dataset \cite{camargo2021comprehensive} and our collected lower-limb kinematic data of five able-bodied subjects. Data collection was approved and performed under the supervision of the Sustech Medical Ethics Committee (approval number: 20210009, date: 2021/3/2). The errors between the estimated gait phase and the actual gait phase were calculated, and the results were compared with two previously proposed gait phase estimation methods: (1) using the pair of $\phi$ and $\int \phi$ and (2) using the pair of $\phi$ and $\dot \phi$. The virtual constraints of the knee and ankle angles were evaluated by calculating the errors between the desired joint angles and the real joint angles of the subjects. The proposed joint angle tracking controller was validated on a powered transfemoral prosthesis.

\newcommand{\ra}[1]{\renewcommand{\arraystretch}{#1}}
\begin{table}[!t]
	\centering
	\caption{Details of eight subjects}%
	\ra{0.9}
	
	\begin{tabular}{cccccc}
		\toprule
		No. & Gender & Height (cm) & Weight (kg) & Age & Source \\\hline
		1& Female & 164 &53 &20&dataset \\
		2& Female & 158 &55&23&dataset\\
		3& Male & 173 &66&25&dataset \\
		4& Female & 168 &54&24&campus\\
		5&Male&170&66&24&campus\\
		6&Male&180&90&25&campus\\
		7&Female&164&54&28&campus\\
		8&Male&183&71&24&campus \\
		\bottomrule
	\end{tabular}
	\label{tab:sub}
\end{table}
\subsection{The lower-limb motion datasets}
In the public dataset \cite{camargo2021comprehensive}, subjects were instructed to walk with five locomotion modes, including : level walking (LW), ramp ascent (RA) and descent (RD), stairs ascent (SA) and descent (SD). The whole-body joint angles of the subjects in the above locomotion modes, including eight transition modes (LW-RA, RA-LW, LW-RD, RD-LW, LW-SA, SA-LW, LW-SD and SD-LW) were recorded in the dataset. The walking data of three subjects from this dataset (refer to subject 1 $\sim$ 3 in Table.~\ref{tab:sub}) were utilized to evaluate the proposed gait phase estimation and smoothing method.

\begin{figure*}[!t]
	\centering
	\subfigure[The IMUs attached to the leg of the subjects]{
		\includegraphics[width=0.38\columnwidth]{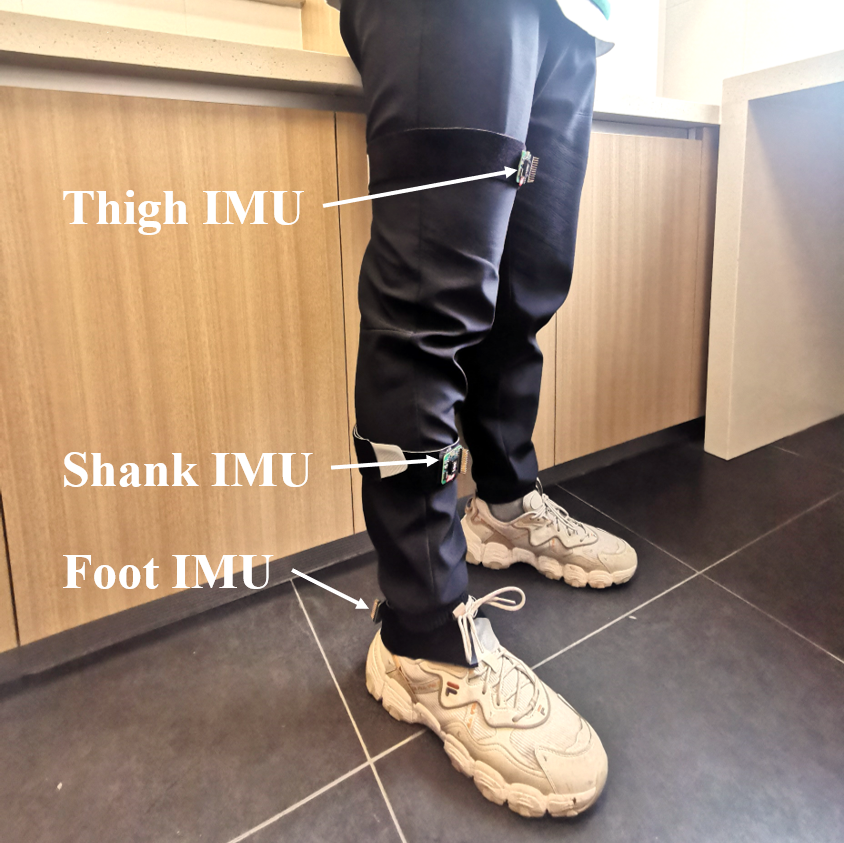}
	}\qquad
	\subfigure[The flat treadmill for LW\_fast and LW\_slow modes]{
		\includegraphics[width=0.38\columnwidth]{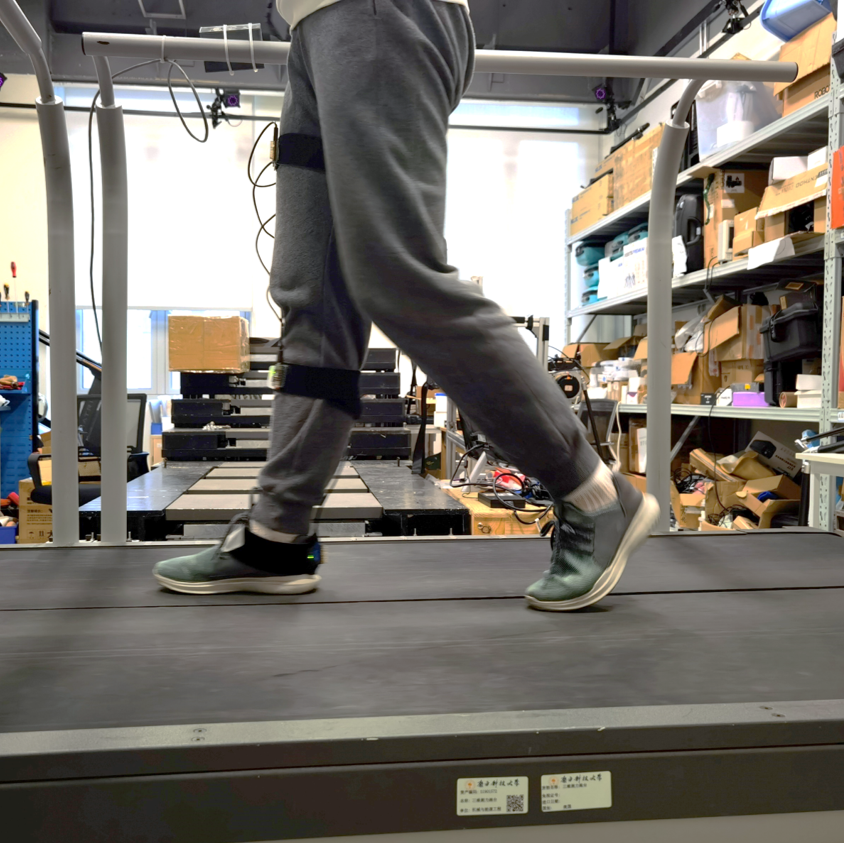}
	}\qquad
	\subfigure[The treadmill with $8^{\circ}$ inclination for RA and RD modes]{
		\includegraphics[width=0.38\columnwidth]{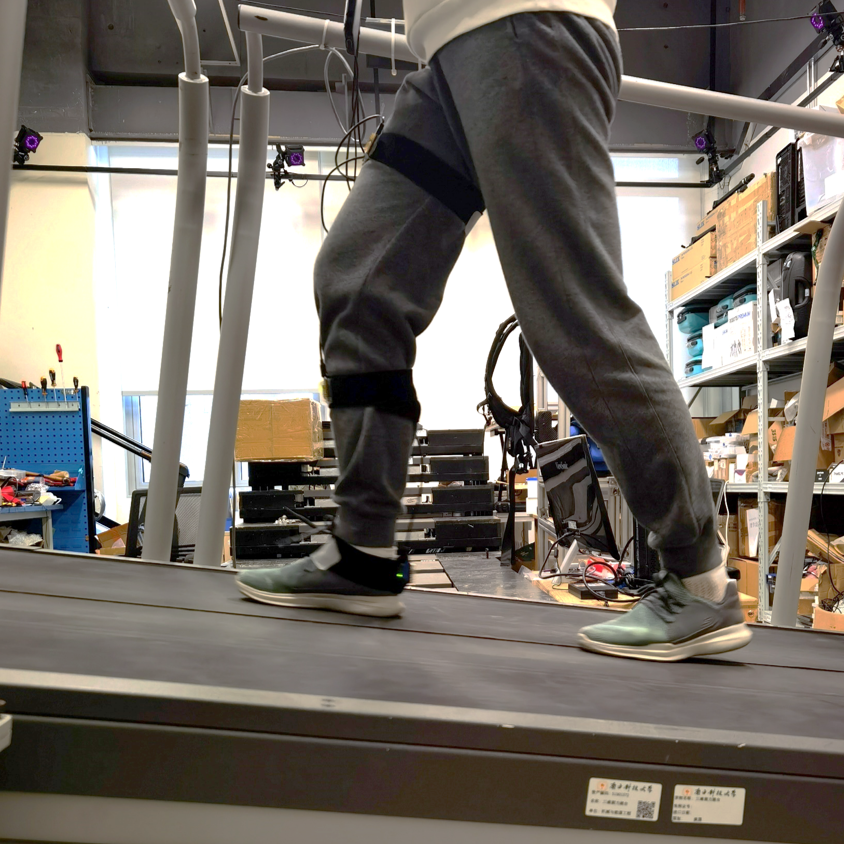}
	}\qquad
	\subfigure[The stairs for SA and SD modes]{
		\includegraphics[width=0.38\columnwidth]{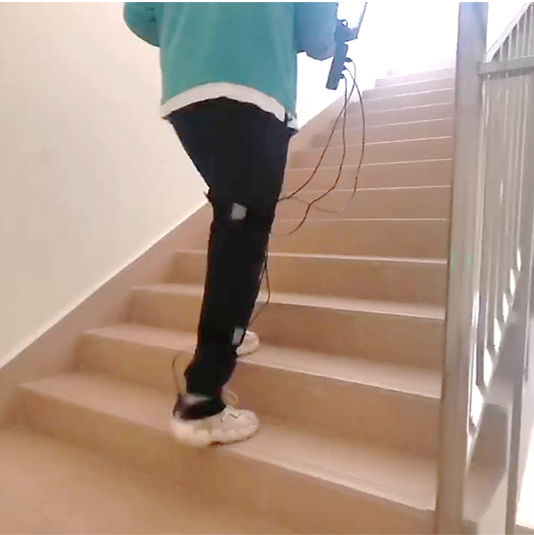}
	}\qquad
	\caption{Lower-limb motion data collection devices and environments. (a) shows the IMUs attached to the legs of the subjects. (b) $\sim$ (d) show the treadmill and the stairs where the data were collected.}
	\label{fig:datacollection}
\end{figure*}

We also recruited five able-bodied subjects (subject 4 $\sim$ 8 in Table.~\ref{tab:sub}) with various heights, weights and genders on campus to walk with the above five locomotion modes and recorded their lower-limb kinematic data. The data collection devices and environments are shown in Fig.~\ref{fig:datacollection}. Three IMUs are attached to the thigh, shank, and heel to record the Euler angles and accelerations every 0.04 s. The subjects were instructed to walk on a flat treadmill with 0.8 m/s and 1.2 m/s (LW\_slow and LW\_fast), an inclined treadmill with 0.8 m/s (RA and RD), and stairs (SA and SD).

\subsection{Offline Evaluation on the Datasets}
The lower-limb motion data was processed to divide time-series data into separate gait cycles. The beginning of the gait cycles is characterized by a large vertical acceleration measured by the foot IMU. \hl{In this paper, the gait phase $s$ maps the progress of the current gait cycle linearly to $[0,100\%]$, and its actual value at time $t$ is calculated by:}
\begin{equation}
	s_t = (t-t_h)/T * 100\%,
	\label{eq:phase_value}
\end{equation}	
\hl{where $t_h$ is the time of the last heel-strike event, $T$ is the duration of the current gait cycle.} The hyperparameters of the proposed gait phase model in Eq.~\eqref{eq:estimate} were fitted from the actual gait phase $s$ and the mean thigh angle profile $\Phi$ in all the gait cycles . The virtual constraint models for the knee and ankle angles in each locomotion mode were also built from the actual gait phase and the mean joint angles. 

The gait phase estimated from the resulting models and the Kalman filtered-based smoother was evaluated by the mean errors (ME), the root mean squared errors (RMSE), and the Pearson correlation coefficients \cite{schober2018correlation} ($r$) between the estimated gait phase $\hat s$, the smoothed gait phase $\tilde{s}$, and the actual gait phase $s$ at all the time steps in all the gait cycles. The virtual constraints for the knee and ankle angles are evaluated by ME, RMSE, and $r$ between the desired angles $\alpha_{\text{d}i} $ determined by the virtual constraints and the actual joint angles $\alpha_i$ ($i \in {1,2}$).

The evaluation results on the public datasets are presented in Table.~\ref{tab:dataset_results}. The estimation accuracy was evaluated for all the five rhythmic locomotion modes and eight transition modes. Despite that the errors of the estimated $\hat s$ are low enough for every locomotion mode, the smoothed gait phase $\tilde{s}$ are even closer to the actual gait phase. This demonstrates the advantages of the Kalman filter-based smoother.

\begin{table*}[!t]
	\centering
	\caption{Performance evaluation results on the public dataset}%
	\ra{0.9}
	\begin{tabular}{ccccccccccccccccc}
		\toprule
		\multirow{2.5}{*}{Mode}  && \multicolumn{3}{c}{$\hat{s}(\%)$} && \multicolumn{3}{c}{$\tilde{s}(\%)$} && \multicolumn{3}{c}{$\alpha_{1}\left({ }^{\circ}\right)$} && \multicolumn{3}{c}{$\alpha_{2}\left({ }^{\circ}\right)$} \\
		\cmidrule{3-5} \cmidrule{7-9} \cmidrule{11-13} \cmidrule{15-17}
		&& Mean & RMSE & $r$ && Mean & RMSE & $r$ && Mean & RMSE & $r$ && Mean & RMSE & $r$ \\
		\midrule
		LW && -0.16& 1.77& 1.00&& -0.22& 1.25& 1.00&& 0.09& 3.05& 0.99&& -0.11& 2.59& 0.94\\
		RA && -0.01& 1.32& 1.00&& 0.00& 0.34& 1.00&& 0.02& 2.38& 0.99&& -0.09& 1.93& 0.98\\
		RD && -0.68& 6.16& 0.98&& -0.67& 2.97& 1.00&& 1.20& 5.53& 0.96&& -0.69& 3.64& 0.94\\
		SA  && -0.00& 1.17& 1.00&& 0.01& 0.43& 1.00&& 0.13& 2.39& 1.00&& -0.05& 2.28& 0.96\\
		SD  && 0.03& 2.21& 1.00&& 0.01& 0.77& 1.00&& 0.04& 3.27& 0.99&& 0.06& 2.54& 0.99\\
		LW-RA  && 0.02& 1.45& 1.00&& -0.01& 0.56& 1.00&& 0.04& 5.13& 0.95&& -0.13& 3.73& 0.94\\
		RA-LW  && -0.02& 2.64& 1.00&& 0.06& 0.74& 1.00&& 0.09& 3.60& 0.98&& -0.08& 2.20& 0.97\\
		LW-RD  && -0.47& 5.83& 0.98&& -0.48& 2.57& 1.00&& 1.06& 6.92& 0.94&& -0.59& 2.94& 0.96\\
		RD-LW  && -0.72& 3.48& 0.99&& -1.04& 2.93& 1.00&& 1.02& 5.81& 0.96&& -0.58& 4.29& 0.95\\
		LW-SA  && 0.00& 1.13& 1.00&& 0.03& 0.58& 1.00&& 0.04& 7.30& 0.96&& -0.04& 4.17& 0.88\\
		SA-LW  && -0.04& 0.85& 1.00&& -0.03& 0.27& 1.00&& 0.02& 2.56& 0.99&& 0.01& 1.70& 0.99\\
		LW-SD  && 0.06& 1.65& 1.00&& 0.07& 0.59& 1.00&& 0.02& 4.62& 0.98&& -0.09& 3.44& 0.98\\
		SD-LW  && -0.02& 3.74& 0.99&& 0.03& 2.54& 1.00&& -0.40& 8.69& 0.94&& -0.20& 4.56& 0.91\\
		\bottomrule
	\end{tabular}
	\label{tab:dataset_results}
\end{table*}

On our collected dataset, the proposed method was compared with two previous gait phase estimation methods, denoted by $\phi$~\&~$\int \phi$ and $\phi$~\&~$\dot \phi$, respectively. \hl{These two methods were chosen for comparison because they were typically and widely used in recent gait phase estimation works}\cite{hong2021phase,quintero2018continuous,8009448,7469796,9350301} \hl{and have been validated through robotic experiments on prostheses, thus can represent the state of the art. Some other references in this paper\cite{Zhang_2022,thatte2019robust} were not considered for comparison because the former one has not been tested on a prosthesis and the latter one was designed only for stance phase.}

\hl{The method $\phi$~\&~$\int \phi$ estimated the gait phase as:}
\begin{equation}
   \hat s= \atantwo((\int\phi+\Gamma) z,(\phi+\gamma))/2\pi,
\end{equation}
\hl{where the parameters $(z,\gamma,\Gamma)$ were calculated for each locomotion modes respectively, using the mean thigh angle profile $\Phi$ derived from the collected data: $z =\frac{\left|\Phi_{\max }-\Phi_{\min }\right|}{\left|\int\Phi_{\max }-\int\Phi_{\min }\right|}$, $\gamma=-\left(\frac{\Phi_{\max }+\Phi_{\min }}{2}\right)$, and $\Gamma=-\left(\frac{\int\Phi_{\max }+\int\Phi_{\min }}{2}\right)$.}

\hl{The method $\phi$~\&~$\dot \phi$ estimated the gait phase as:}
\begin{equation}
	\hat s= \atantwo((\dot\phi+\Lambda) k,(\phi+\lambda))/2\pi,
\end{equation}
\hl{where the parameters were also calculated for each locomotion modes respectively: $k =\frac{\left|\Phi_{\max }-\Phi_{\min }\right|}{\left|\dot\Phi_{\max }-\dot\Phi_{\min }\right|}$, $\lambda=-\left(\frac{\Phi_{\max }+\Phi_{\min }}{2}\right)$, and $\Lambda=-\left(\frac{\dot\Phi_{\max }+\dot\Phi_{\min }}{2}\right)$. The above equations calculating the parameters of the two methods were all from the cited references.} 

\begin{figure*}[!t]
	\centering
	\includegraphics[width = 1.8\columnwidth]{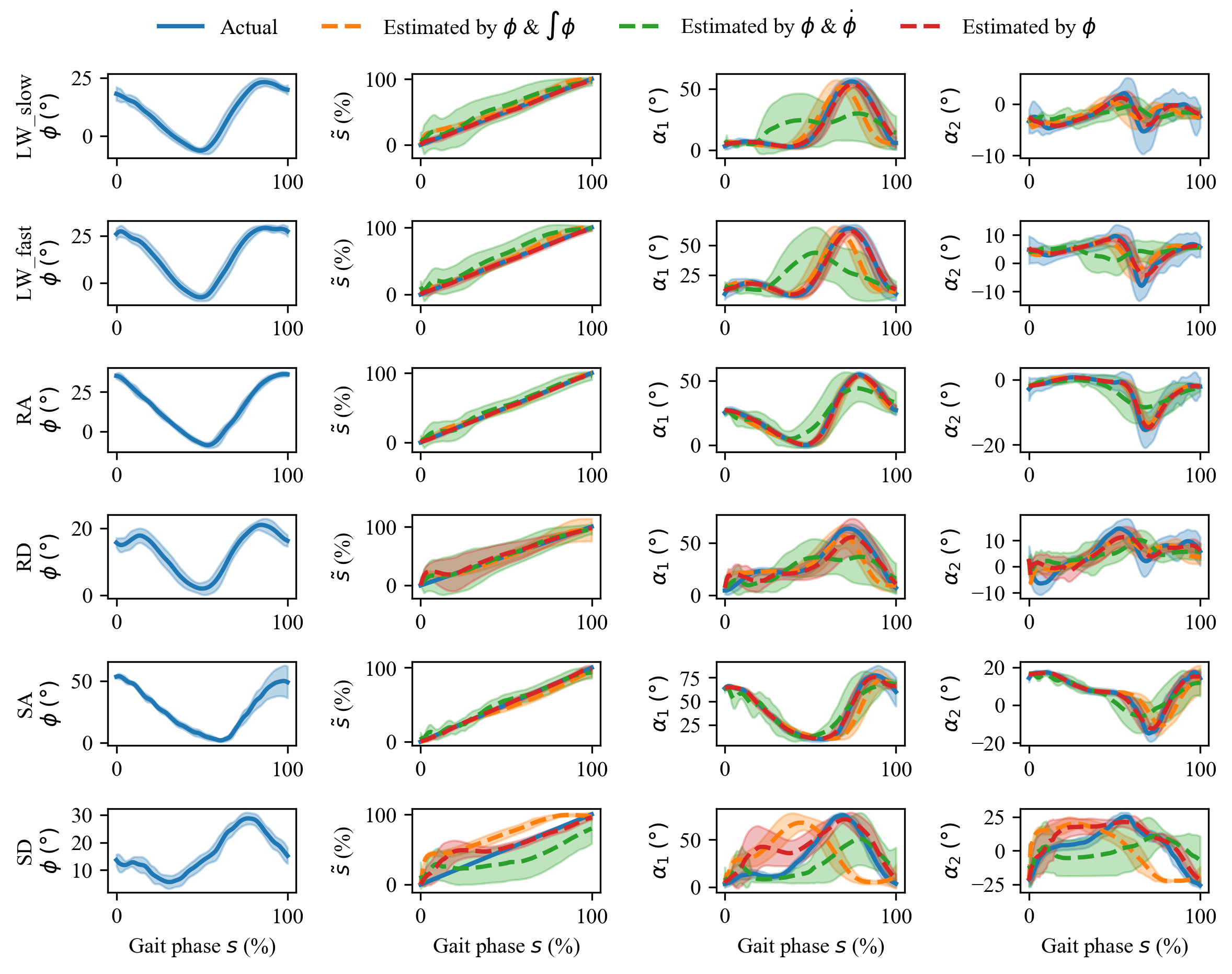}
	\caption{The gait phase and joint angle of subject 8 estimated by our proposed method and two other methods. The solid blue curves and the shaded region are the mean and standard deviation of actual values. The orange and the green curves and the shaded region are the estimation results of the two benchmark methods. The red curves and the shaded region are the results of our proposed method. }
	\label{fig:curves}
\end{figure*}

The comparison results are presented in Table.~\ref{tab:our_results}. It can be seen that our proposed method can achieve lower estimation errors for the gait phase and also the desired joint angles. To better illustrate the gait phase and joint angle estimation results with our proposed method and two other methods, the results of subject 8 were presented in Fig.~\ref{fig:curves}. It can be seen that the gait phase curves and the joint angle curves estimated from the proposed method are closer to the actual values compared with the other two methods. The estimation performance of $\phi$~\&~$\dot \phi$ is the worst, because the thigh angle velocity can be very noisy and less stable than the thigh angle itself.

\begin{table*}[!t]
	\centering
	\caption{Performance evaluation results on our collected dataset}%
	\ra{0.9}
	\begin{tabular}{cclcccccccccccc}
		\toprule
		\multirow{2.5}{*}{ Mode }&& \multirow{2.5}{*}{ Method }&&\multicolumn{3}{c}{$\tilde{s}(\%)$} && \multicolumn{3}{c}{$\alpha_{1}\left({ }^{\circ}\right)$} && \multicolumn{3}{c}{$\alpha_{2}\left({ }^{\circ}\right)$} \\
		\cmidrule{5-7} \cmidrule{9-11} \cmidrule{13-15}
		&  &&& Mean & RMSE & $r$ && Mean & RMSE & $r$ & & Mean & RMSE & $r$ \\
		\midrule \multirow{3}{*}{ LW\_slow }& & $\phi$ \& $\int \phi$ && 2.55& 7.12& 0.97&& -2.61& 10.52& 0.87&& 0.14& 4.96& 0.99\\
		&& $\phi$ \& $\dot{\phi}$ && 9.18& 20.32& 0.85&& -0.00& 25.58& 0.25&& -0.98& 7.26& 0.98\\
		&& Ours: $\phi$& & -0.46& 4.44& 0.99&& -0.17& 5.61& 0.96&& 0.10& 3.95& 0.99\\
		\hline \multirow{3}{*}{ LW\_fast } && $\phi$ \& $\int \phi$&& 1.62& 5.87& 0.98&& -2.77& 11.05& 0.87&& 0.28& 5.62& 0.99\\
		&& $\phi$ \& $\dot{\phi}$ && 5.77& 15.89& 0.90&& -2.92& 24.46& 0.37&& -0.61& 9.09& 0.97\\
		&& Ours: $\phi$ && -0.41& 4.78& 0.99&& -0.26& 6.2& 0.96&& -0.05& 4.58& 0.99\\
		\hline \multirow{3}{*}{ RA } && $\phi$ \& $\int \phi$&& -1.52& 4.42& 0.99&& -1.35& 6.99& 0.95&& 0.49& 5.21& 0.98\\
		&& $\phi$ \& $\dot{\phi}$ && 3.1& 13.36& 0.91&& 0.58& 12.58& 0.85&& -0.44& 6.86& 0.97\\
		&& Ours: $\phi$ && -0.23& 2.85& 1.0&& -0.22& 4.38& 0.98&& 0.02& 3.85& 0.99\\
		\hline \multirow{3}{*}{ RD }&& $\phi$ \& $\int \phi$ && 4.45& 11.84& 0.93&& -3.58& 16.15& 0.76&& 0.16& 6.3& 0.98\\
		&& $\phi$ \& $\dot{\phi}$ && 8.11& 23.19& 0.8&& -6.61& 25.21& 0.48&& -1.01& 7.62& 0.97\\
		&& Ours: $\phi$ && 3.46& 18.27& 0.83&& -2.8& 13.59& 0.84&& 0.08& 5.9& 0.98\\
		\hline \multirow{3}{*}{ SA }&& $\phi$ \& $\int \phi$ && -3.63& 9.97& 0.95&& -2.87& 12.6& 0.88&& 0.23& 6.89& 0.76\\
		&& $\phi$ \& $\dot{\phi}$ && 2.37& 17.34& 0.84&& 1.29& 17.84& 0.75&& -0.68& 8.44& 0.65\\
		&& Ours: $\phi$ && 0.75& 10.04& 0.94&& -0.21& 7.64& 0.96&& -0.14& 5.44& 0.86\\
		\hline \multirow{3}{*}{ SD }&& $\phi$ \& $\int \phi$ && 19.49& 24.25& 0.88&& 5.53& 33.55& 0.26&& -0.99& 17.88& 0.46\\
		&& $\phi$ \& $\dot{\phi}$ && -16.66& 30.56& 0.58&& -8.51& 32.44& 0.19&& -1.84& 22.04& 0.05\\
		&& Ours: $\phi$&& 5.14& 17.73& 0.82&& 9.59& 22.19& 0.71&& 1.99& 11.2& 0.76\\
		\bottomrule
	\end{tabular}
	\label{tab:our_results}
\end{table*}

\subsection{Real-time Evaluation on the Prosthesis}

\begin{figure}[!t]
	\centering
	\includegraphics[width = 0.9\columnwidth]{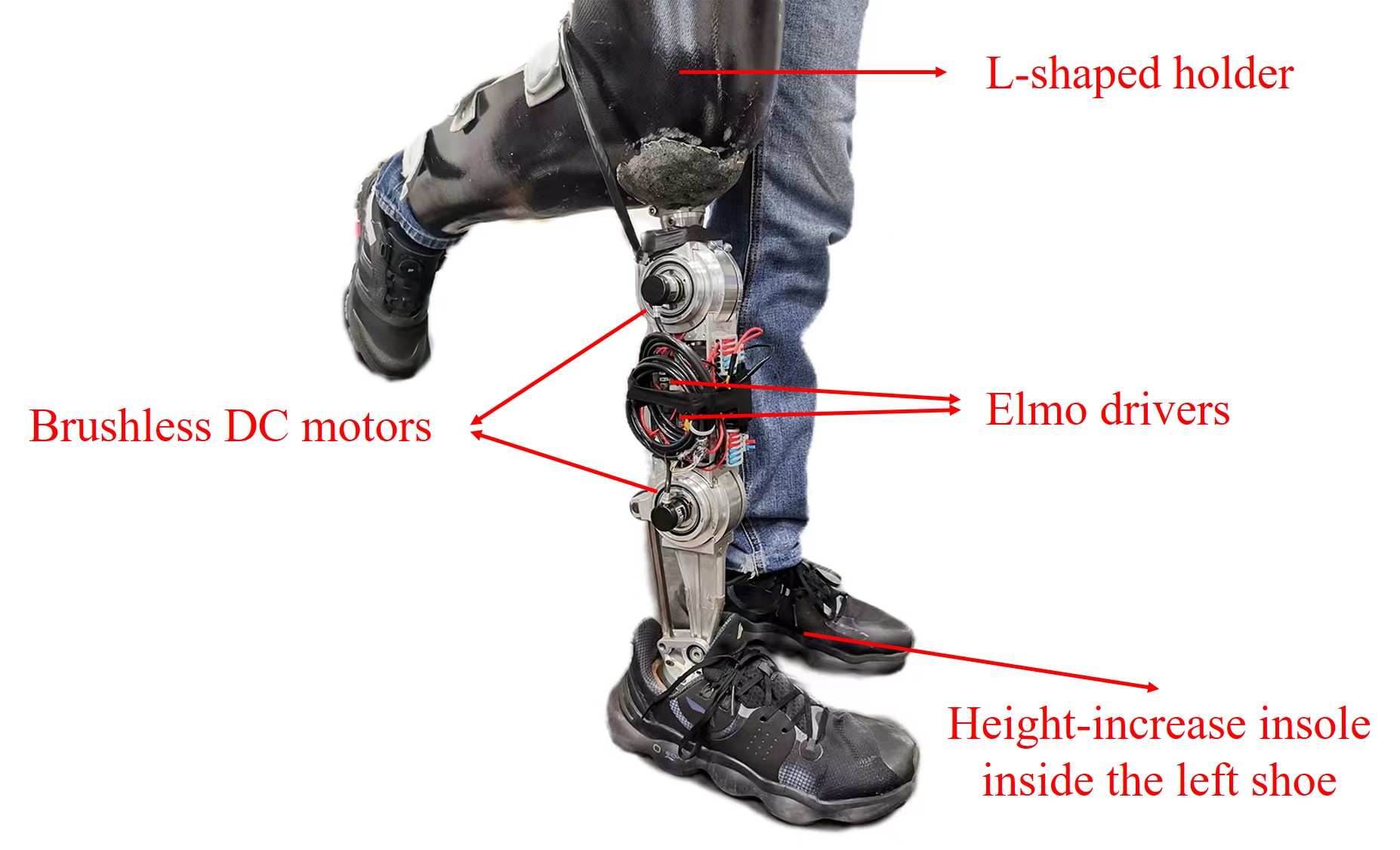}
	\caption{The powered transfemoral prosthesis. Two brushless DC electric motors actuated by the Elmo drivers serve as the knee and the revolute pair of the ankle. The able-bodied subject wears the prosthesis with an L-shaped holder. A height-increase insole was used in the left shoe of the subject to compensate for the difference between the lengths of his leg and the prosthesis.}
	\label{fig:prosthesis}
\end{figure}

In order to validate the proposed gait phase estimation method and the gait phase-based controller, an experiment was conducted on a powered transfemoral prosthesis shown in Fig.~\ref{fig:prosthesis}. Subject 8 was instructed to wear the prosthesis with an L-shaped holder and walk on the treadmill with 0.8 m/s speed. The process of the experiment was recorded and presented in the supplementary video of this paper. An IMU was attached to the L-shaped holder to measure the thigh angle of the subject. The proposed gait phase estimation and smoothing method was applied to estimate the gait phase in real time. The prosthesis was driven by the proposed controller to track the joint angles determined by the virtual constraints. The current commands $\tau$ calculated with Eq.~\eqref{eq:controller} were sent to the motor driver every 0.03 s. The knee angle and the ankle angle of the prosthesis were measured by the encoders of the motors. 

\begin{figure*}[!t]
	\centering
	\includegraphics[width = 2.1\columnwidth]{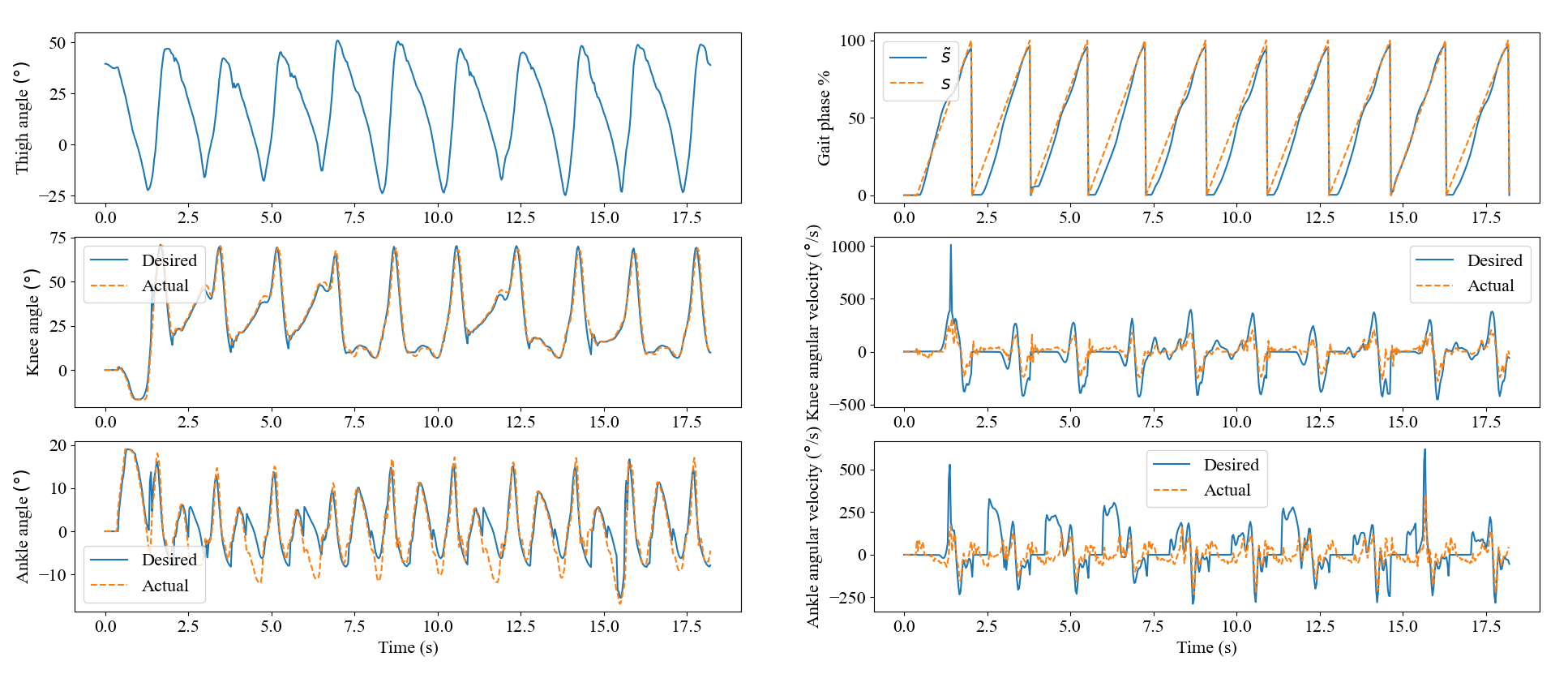}
	\caption{An example series of recorded data and the smoothed estimated gait phase in the experiment. The desired joint angles are determined by the virtual constraint models. The actual thigh angle is measured by an IMU. The actual knee and ankle angle of the prosthesis is measured by the encoders.}
	\label{fig:exp}
\end{figure*}

An example series of thigh angle and the real-time estimated gait phase and the joint angles are presented in Fig.~\ref{fig:exp}. \hl{The ground truth of $s$ in the experiment is calculated by Eq.~\eqref{eq:phase_value}.} It can be seen that the smoothed estimated gait phase $\tilde{s}$ is highly aligned to the actual gait phase. The RMSE of $\tilde{s}$ is $7.50\%$, and the correlation coefficient between $\tilde{s}$ and $s$ is 0.99. The knee angle and the ankle angle of the prosthesis also show good rhythmicity and follow the subject's motion pattern in the LW\_slow locomotion mode.

\subsection{Discussions}
From the above results, the superiority of the proposed gait phase estimation method can be proved. The feasibility of the proposed gait phase-based controller has also been validated.

In the offline evaluation of the proposed method, the proposed method can achieve high estimation accuracy in various locomotion modes, including transition modes, which have never been analyzed in other gait phase estimation papers. The gait phase estimation errors can reach 37.6\% lower than other gait phase estimation methods. This result can inspire the researchers to investigate how the gait phase-based controllers can drive the walking-aid robots in complex environments, e.g., outdoor environments where the locomotion modes often change.

It should be noted that in the evaluation results on our collected dataset, the gait phase estimation accuracy in the RD and SD modes was worse than in other modes. It can be explained that human is in an unstable state going down the ramp and the stairs, and the joint angles can vary a lot during the single support stage \cite{zachazewski1993biomechanical,boyaninska2018balance}. This is also a reason why most accidental falls happen during ramp descent and stair descent.

From the experiment on the thigh prosthesis, it can be seen that the estimated gait phase matched the actual gait phase well, however, gait phase estimation results may be further improved, especially at the early stance stage in each gait cycle. The Kalman filter-based smoother designed in this paper is initialized at every heel-strike event, thus leading to inaccurate gait phase changing velocity estimation at the early stance stage. The smoother can be improved by estimating a moving average of the gait phase changing velocity in previous steps, and using the estimated gait phase changing velocity to update the Kalman filter.

\section{Conclusions and Outlooks}
\label{sec:IV}
In this paper, a unified form of piecewise monotonic models was proposed to reveal the relationship between the gait phase and the thigh angle in various locomotion modes. The inverse function of the model was used to estimate the gait phase with only the thigh angle. The estimated gait phase was further smoothed by a Kalman filter-based smoother to mitigate estimation errors and mutations. A joint angle tracking control method was proposed for a transfemoral prosthesis.

The proposed gait phase estimation method was evaluated on a public dataset and our collected dataset and showed high estimation accuracy. Because the measured thigh angle is more stable and less noisy compared with its integral and differential, the proposed method using only the thigh angle can achieve better performance than two existing gait phase estimation methods using the pair of $\phi$ and $\int \phi$ and using the pair of $\phi$ and $\dot \phi$, respectively.

The proposed gait phase-based impedance controller showed good real-time performance in the experiment. But it was only validated in a continuous and rhythmic locomotion mode, further evaluations should be conducted for various locomotion modes, especially the RD mode, the SD mode, and the transition modes. To improve the performance of the powered prosthesis in complex environments, vision-based environment perception can contribute to updating the parameters of the gait phase estimation model to adapt to various types of terrains.

\section{ACKNOWLEDGMENT}
The authors would like to thank Dr. Kuangen Zhang from the University of British Columbia for his contribution to the mathematical reasoning of the piecewise monotonic gait phase model during his visit at Southern University of Science and Technology from 2020 to 2021.

\addcontentsline{toc}{chapter}{Bibliography}

\bibliography{ref}
\bibliographystyle{IEEEtran}

\vfill

\end{document}